\documentclass[conference]{IEEEtran}
\IEEEoverridecommandlockouts

\usepackage{cite}
\usepackage{amsmath,amssymb,amsfonts,subfigure}
\usepackage{algorithmic}
\usepackage{graphicx}
\usepackage{textcomp}
\usepackage{xcolor}
\def\BibTeX{{\rm B\kern-.05em{\sc i\kern-.025em b}\kern-.08em
    T\kern-.1667em\lower.7ex\hbox{E}\kern-.125emX}}

\usepackage{color}
\usepackage{mathtools}
\usepackage{booktabs}

\usepackage{comment}

\newtheorem{lemma}{Lemma}

\newenvironment{Proof}[1]{\medskip\par\noindent{\bf Proof:\,}\,#1}{{\mbox{\,$\blacksquare$}\par}}
\IEEEoverridecommandlockouts
\allowdisplaybreaks

\begin{document}

\title{Balancing Information Accuracy and Response Timeliness in Networked LLMs\\
\thanks{Research of MB was partially supported by BİL2: BILKENT University-TUBITAK BILGEM Consultancy Call for Research EDGE-4-IoT and TUBITAK 2232-B  Fellowship (Project No:124C533).}
}

\author{\IEEEauthorblockN{Yigit Turkmen}
\IEEEauthorblockA{\textit{Dept. of Electrical and Electronics Eng.} \\
\textit{Bilkent University}\\
Ankara, Turkey \\
yigit.turkmen@ug.bilkent.edu.tr}
\and
\IEEEauthorblockN{Baturalp Buyukates}
\IEEEauthorblockA{\textit{School of Computer Science} \\
\textit{University of Birmingham}\\
Birmingham, UK \\
b.buyukates@bham.ac.uk}
\and
\IEEEauthorblockN{Melih Bastopcu}
\IEEEauthorblockA{\textit{Dept. of Electrical and Electronics Eng.} \\
\textit{Bilkent University}\\
Ankara, Turkey \\
bastopcu@bilkent.edu.tr}
}

\maketitle

\begin{abstract}
Recent advancements in Large Language Models (LLMs) have transformed many fields including scientific discovery, content generation, biomedical text mining, and educational technology. However, the substantial requirements for training data, computational resources, and energy consumption pose significant challenges for their practical deployment. A promising alternative is to leverage smaller, specialized language models and aggregate their outputs to improve overall response quality. In this work, we investigate a networked LLM system composed of multiple users, a central task processor, and clusters of topic-specialized LLMs. Each user submits categorical binary (true/false) queries, which are routed by the task processor to a selected cluster of $m$ LLMs. After gathering individual responses, the processor returns a final aggregated answer to the user. We characterize both the information accuracy and response timeliness in this setting, and formulate a joint optimization problem to balance these two competing objectives. Our extensive simulations demonstrate that the aggregated responses consistently achieve higher accuracy than those of individual LLMs. Notably, this improvement is more significant when the participating LLMs exhibit similar standalone performance. 
\end{abstract}


\section{Introduction}
Recent advancements in Large Language Models (LLMs) have revolutionized numerous domains including natural language processing, content generation, and information retrieval. With capabilities ranging from answering complex queries to generating creative content, LLMs like GPT-4, Claude 3 Opus, and LLaMA 4 have demonstrated remarkable performance across a wide spectrum of tasks \cite{brown2020language, touvron2023llama}. These models have been rapidly deployed in various applications, including virtual assistants, content recommendation systems, and automated customer service platforms. With increasing demand for these models in real-world applications, efficient task routing systems that can handle multiple users and optimize the utilization of LLM resources have become paramount.

Despite their capabilities, the deployment of LLMs in production environments presents several technical challenges. First, these models require substantial computational resources, making them costly to operate at scale. Second, their response quality can vary significantly across different types of queries. Third, latency is a crucial factor for user-facing applications, requiring efficient scheduling policies. To mitigate these challenges, recent efforts have focused on developing and fine-tuning smaller LLMs such as LLaMA~3.1-8B \cite{grattafiori2024llama3herdmodels} and Qwen3-8B\cite{yang2025qwen3technicalreport} for improved task-specific accuracy while being resource-efficient. In the near future, these compact models may be deployed directly on personal devices, vehicles, and edge computing systems. Although these personalized, smaller models demonstrate strong performance on targeted tasks, their effectiveness may still be limited when addressing unfamiliar or more diverse queries. To improve reliability across a broader range of tasks, a promising approach is routing user queries to a network of specialized expert LLMs and aggregating their outputs to form more accurate and comprehensive final responses. This raises a central research question that we explore in this work:

\textit{Can we design networked LLM systems that ensure both information accuracy and response timeliness?}     

\begin{figure}[!t]
\centering
\includegraphics[width=\columnwidth]{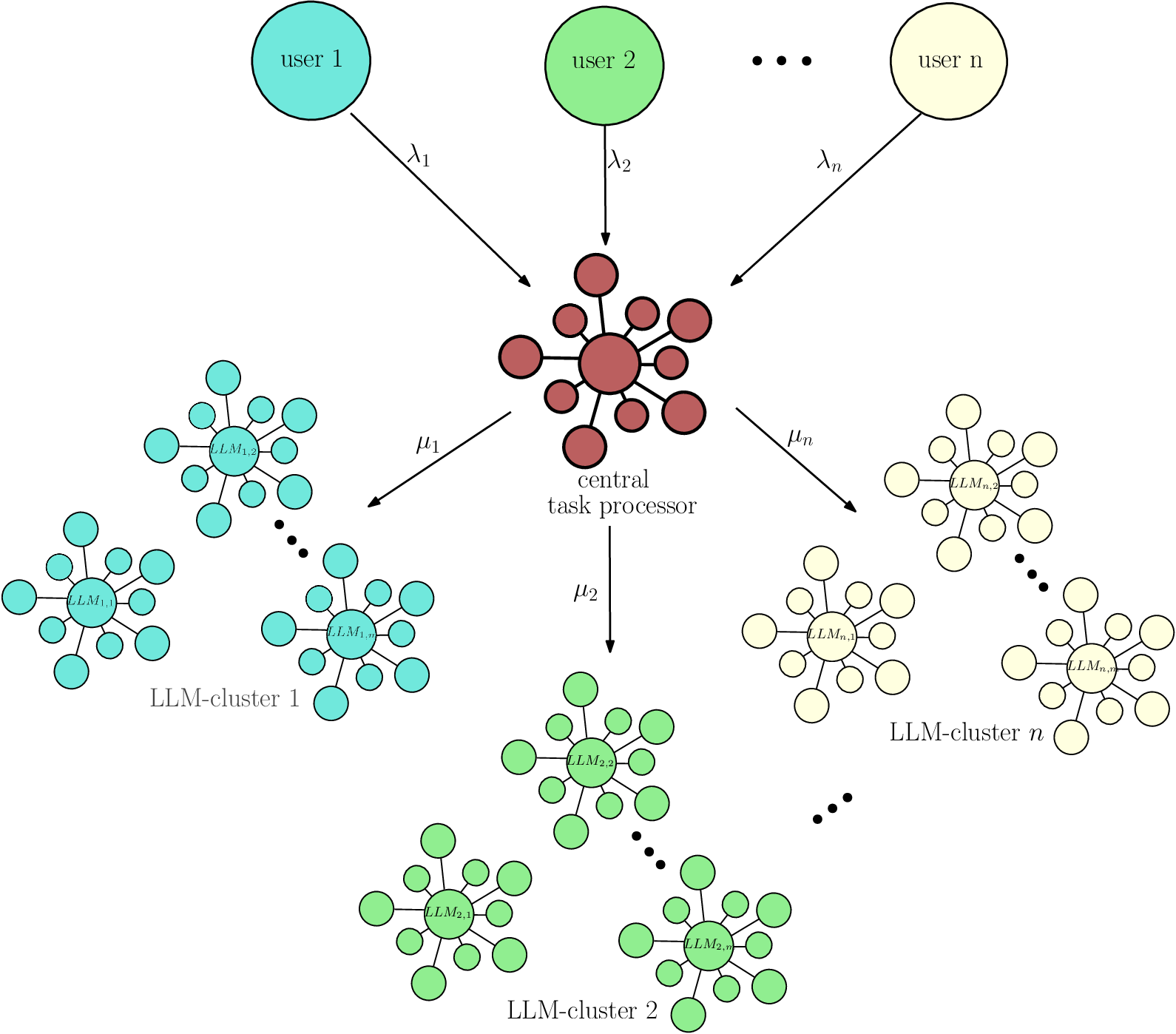}
\caption{Multi-user task routing system architecture with a central task processor connecting $n$ users to multiple task-specialized expert LLM-clusters.}
\label{fig:system}
\end{figure}

To the best of our knowledge, no prior work has developed a formal, analytical model to optimize the trade-off between response accuracy and system timeliness in a multi-LLM architecture. The integration of multiple LLMs in networked environments is a key step towards EGI (Edge General Intelligence), but it presents significant challenges in terms of architecture, accuracy, and arrangement \cite{luo2025toward}. Further, a recent systematic analysis by reference \cite{cemri2025multiagentllmsystemsfail} reveals that many multi-agent systems exhibit high failure rates due to inaccurate system design and agent misalignment, necessitating more principled architectural designs. In this work, we address this challenge by proposing an analytical framework to guide the design of a timely and accurate multi-agent system.

\subsection{Main Contributions}

We consider a networked LLM system where users generate distinct types of binary queries (yes/no or true/false questions) and send them to a central task processor as shown in Fig.~\ref{fig:system}. Having access to multiple clusters of LLMs, the task processor forwards the user queries to a suitable cluster, consisting of $m$ specialized LLMs. After collecting responses from the LLMs within the selected cluster, the task processor generates a final response for the user. Our goal is to determine the number of LLMs, $m$, to query within each cluster to ensure both information accuracy and response timeliness.

Our main contributions are as follows:
\begin{itemize}
    \item Unlike prior work on Mixture-of-Agents (MoA), we focus on the binary query setting and derive a closed-form expression, along with a precise approximation, for the information accuracy of the final aggregated response, under certain assumptions about the system model.  
    \item Leveraging the maximum a posteriori (MAP) estimator, we show that the final aggregated response follows an adaptive majority rule, where the decision threshold is adjusted according to the apriori accuracy of the queries and the expertise levels of the LLMs. 
    \item We measure \textit{timeliness} as the duration between two consecutive \textit{correct answers} returned to the same user. Based on this, we formulate an optimization problem that balances information accuracy with response timeliness.
    \item Through extensive evaluations utilizing various off-the-shelf pre-trained LLMs, we observe a consistent improvement in the information accuracy of the final aggregated response across various question answering (QA) benchmarks. This improvement becomes more notable when the individual LLMs exhibit similar levels of accuracy.       
    
\end{itemize}

\subsection{Related Work}
In this subsection, we review the existing literature on multi-agent LLM systems and efficient query routing. 
\subsubsection{Multi-Agent Collaboration and Architectures}
The foundational MoA framework \cite{wang2024mixture} introduced a layered architecture where agents in subsequent layers refine their responses based on the outputs of the previous layer. Building on this, frameworks like SMoA \cite{li2025smoa}
and RMoA \cite{xie2025rmoaoptimizingmixtureofagentsdiversity} have introduced sparsity and residual connections to improve the efficiency and robustness of this iterative process. Another paradigm involves multi-agent debate systems \cite{du2023improving, liang2023encouraging}, which use a dialectical process in which agents critique each other's reasoning to arrive at a more accurate answer. However, the iterative, multi-round nature of these frameworks makes them unsuitable for applications where predictable, low latency responses are important. In contrast, we introduce a non-iterative architecture explicitly designed for timeliness, i.e., low latency. While our model assumes homogeneity within an agent cluster for analytical tractability, i.e., each cluster consists of multiple LLM agents that share the same accuracy and processing time while making independent decisions, works like X-MAS \cite{ye2025xmasbuildingmultiagentsystems} demonstrate the performance benefits of building systems with heterogeneous LLMs.

\subsubsection{Efficient Query Routing}
Alternatively, some studies have focused on managing resources efficiently, through query routing. Some frameworks, like FrugalGPT \cite{chen2023frugalgptuselargelanguage}, employ a sequential LLM cascade that can introduce unpredictable latency. To mitigate this, predictive \textit{``route-to-one"} systems aim to select the best model for a query in a single step. These include dynamic, learning based approaches like MixLLM \cite{wang-etal-2025-mixllm}, methods that utilize uncertainty estimation such as the Confidence-Driven LLM Router \cite{zhang2025leveraginguncertaintyestimationefficient}, and adaptive techniques like LightRouter\cite{zhang2025lightrouterefficientllmcollaboration} which makes a selection after a few boot tokens. While these works offer promising and sophisticated methods for selecting a single best agent, some studies suggest that routing to a single model is not always ideal\cite{srivatsa2024harnessingpowermultipleminds,ding2024hybridllmcostefficientqualityaware}. Our work diverges by exploring a \textit{``route-to-many aggregation"} approach. Our primary contribution is a mathematical model that determines \textit{how many} agents to query, rather than simply selecting the best single agent. This focus on optimizing the system level behavior for timeliness also aligns with the goals of \cite{mitra2024distributed}, which provides a queueing stability analysis to ensure bounded latency in an agent network.

The remainder of this paper is organized as follows:
In Section~\ref{Sec:System_model}, we present the system model, derive analytical expressions for information accuracy and response timeliness, and formulate an optimization problem that balances these two objectives.
Section~\ref{Sec:Opt_of_LLMs} analyzes how the objective function varies with the number of LLMs in each cluster and introduces our solution approach to determine a sub-optimal number of LLMs for the optimization problem at hand.
In Section~\ref{Sec:num_result}, we provide detailed simulation results using various off-the-shelf pre-trained LLMs to demonstrate the effectiveness of our approach.
Finally, Section~\ref{sec:conc} concludes the paper with a discussion and outlines potential directions for future research.            

\section{System Model and Problem Formulation}\label{Sec:System_model}
In this work, we consider a system consisting of $n$ users generating binary (true/false) queries, a central task processor, i.e., router, and a set of LLMs specialized to handle specific types of user queries, as illustrated in Fig.~\ref{fig:system}. 

Each user $i$ generates binary queries according to a Poisson process with rate $\lambda_i$. The prior probability that a query from user $i$ is true is denoted by $w_i \in [0,1]$, which is known to the task router. Queries from each user belong to different categories and must be processed by appropriately specialized LLMs. We assume that the task processor is aware of the set of $m$ LLMs that are capable of answering queries from user $i$. We refer to this set of $m$ LLMs as the \textit{LLM-cluster-$i$} (or simply, cluster $i$), and denote each of its members as $\{LLM_{i,1},\cdots,LLM_{i,m}\}$. Upon receiving a query from user $i$, the task processor routes it to the corresponding LLM cluster $i$, as we assume that user query categories are distinct.\footnote{In practice, the number of users and clusters need not be equal; a single cluster may serve all users whose queries fall into the same category. In our setting, we represent each query category/group with a single user.}

An example such scenario is a patient triage chatbot, where users interact with a triage assistant that routes cases to specialist models for diagnosis. That is, each user’s query is routed to a cluster of LLMs relevant to their symptoms. For instance, if a user asks ``Is it likely that I have COVID-19 based on my symptoms?'', their query would be sent to the cluster with LLMs specialized in infectious diseases.

We assume that the task processor can only receive tasks while it is idle. All the queries that arrive while the task processor is busy, i.e., while responding  to other users, are dropped. When a query from user 
$i$ is accepted, the task processor forwards it to all LLMs in cluster $i$. The transmission time of a query from the task processor to each LLM in cluster~$i$ is modeled as an exponentially distributed random variable with rate~$\mu_i$. After receiving the query, each LLM in cluster $i$ requires a fixed processing time $t_i$ to generate a response. Each LLM in cluster~$i$ returns the correct answer with probability~$p_i$. That is, each LLM in cluster $i$ responds to the query with the correct label (true or false) with probability~$p_i \in [0,1]$. In this work, we focus on a setting where all LLMs in cluster~$i$ share the same processing time~$t_i$ and the same correctness probability~$p_i$.\footnote{In the general system, both the success probabilities~$p_i$ and processing times~$t_i$ may vary across the LLMs within a cluster. We adopt the assumption of identical $p_i$ and $t_i$ for all LLMs in cluster~$i$ to enable analytically tractable solutions that capture the trade-off between accuracy and timeliness. In our experiments, where we simulate this system, we use LLMs whose success probabilities and response times vary around~$p_i$ and~$t_i$, respectively.} We assume that the LLM responses are independent and are returned instantaneously to the task processor.  The task processor aggregates these responses to generate the final response for the users. 

Next, we evaluate the information accuracy in this system.

\subsection{Information Accuracy}
In this subsection, we explore how to aggregate the responses from the $m$ LLMs to produce a final response at the task processor, leveraging a distributed fact-checking mechanism to enhance reliability, as proposed in~\cite{Touri23_ISIT}. For a query from user~$i$, let the response from $LLM_{i,j}$ be denoted by $R_{i,j}$ which is a binary random variable taking values -1 or 1.
We denote the query of type~$i$ being true as $U_i = 1$ and false as $U_i = -1$, with the prior probability $P(U_i = 1) = w_i$ and $P(U_i = -1) = 1 - w_i$.
 Similarly, as mentioned above, we have $P(R_{i,j} = s| U_i = s) = 1-P(R_{i,j} = -s| U_i = s) = p_i$ for $s\in\{-1,1\}$. We denote the set of all realized responses generated by the LLMs in cluster $i$ as $ \boldsymbol{r_i} = \{r_{i,1},\cdots, r_{i,m}\}$.\footnote{In other words, $LLM_{i,j}$'s response is modeled as a binary random variable $R_{i,j}$ and its realization is $r_{i,j}$. In vector form, we have $\boldsymbol{R_i} = \boldsymbol{r_i}$.} Our goal is to design the MAP estimator such that the probability of making the incorrect estimation is minimized. Then this MAP estimator \cite{Touri23_ISIT} is given by
\begin{align}
    \hat{U}_{MAP}(\boldsymbol{r_i}) = \begin{cases} 
      1 & \text{if } \frac{P(U_i =1|\boldsymbol{R_i} = \boldsymbol{r_i})}{P(U_i =-1|\boldsymbol{R_i} = \boldsymbol{r_i})} \geq 1\\
      -1 &  \text{if } \frac{P(U_i =1|\boldsymbol{R_i} = \boldsymbol{r_i})}{P(U_i =-1|\boldsymbol{R_i} = \boldsymbol{r_i})} < 1 
   \end{cases}.
\end{align}
Let $\mathcal{Q}_i$ denote the set of responses equal to 1, with cardinality $k_i$, i.e., $card(\mathcal{Q}_i) = k_i$. Then, we have 
\begin{align*}
    \frac{P(U_i =1|\boldsymbol{R_i} = \boldsymbol{r_i})}{P(U_i =-1|\boldsymbol{R_i} = \boldsymbol{r_i})}  = \frac{w_i \prod_{j\in \mathcal{Q}_i } p_i  \prod_{j\in \mathcal{Q}_i^c } (1-p_i) }{(1-w_i) \prod_{j\in \mathcal{Q}_i } (1-p_i)  \prod_{j\in \mathcal{Q}_i^c } p_i}.
\end{align*}
Since all $LLM_{i,j}$ have the same success probability $p_i$, the expression above can be rewritten as: 
\begin{align}\label{exp-M}
    \frac{P(U_i =1|\boldsymbol{R_i} = \boldsymbol{r_i})}{P(U_i =-1|\boldsymbol{R_i} = \boldsymbol{r_i})}  = \frac{w_i}{1-w_i}\left(\frac{ p_i }{ 1-p_i}\right)^{2k_i-m}.
\end{align}
Thus, the aggregated response is \textit{`correct',} i.e., $\hat{U}_{MAP} =1$, when $\frac{w_i}{1-w_i}\left(\frac{ p_i }{ 1-p_i}\right)^{2k_i-m} \geq 1$ such that for $p_i\geq 0.5$ we have
\begin{align}\label{eqn:optimum_k_i}
    k_i \geq  \frac{m}{2} + \frac{\log\left(\frac{1-w_i}{w_i}\right)}{2\log\left(\frac{1-p_i}{p_i}\right)  } = k_i^*. 
\end{align}
Then, for $p_i\geq 0.5$, the optimum MAP estimator becomes  
\begin{align}\label{eqn:optimum_MAP_rule}
    \hat{U}_{MAP}(\boldsymbol{r_i}) = \begin{cases} 
      1,& \text{if } k_i  \geq k_i^*\\
      -1, &  \text{otherwise,}
   \end{cases}
\end{align}
where $k_i^*$ is provided in (\ref{eqn:optimum_k_i}).\footnote{Note that if $p_i< 0.5$, the optimal MAP estimator becomes $\hat{U}_{MAP}(\boldsymbol{r_i}) = -1$ if  $k_i  \geq k_i^*$ and 1, otherwise. Thus, the MAP estimator works for the entire success probability range of individual LLMs, that is, $0\leq p_i\leq 1$. }
Note that when $w_i = 0.5$, we have $k_i^* = \frac{m}{2}$ meaning the optimal MAP estimator in (\ref{eqn:optimum_k_i}) reduces to a simple majority rule. If $w_i > 0.5$, then $k_i^*<\frac{m}{2}$ indicating that as the prior probability of a query being true increases, fewer confirmations from the LLMs are needed to infer a true outcome. Conversely, if $w_i < 0.5$, more confirmations are required, i.e., $k_i^*>\frac{m}{2}$. A similar reasoning applies to $p_i$: when $p_i>0.5$, higher LLM accuracy lowers the required number of positive assertions, i.e., $k_i^*<\frac{m}{2}$. Thus, the optimal $k_i^*$ acts as an adjusted majority rule, where the adjustment accounts for both the prior probability $w_i$ and the individual LLMs' accuracy $p_i$. 

With the MAP estimator in~(\ref{eqn:optimum_MAP_rule}), the accuracy of the aggregated response generated by the task processor, denoted by $p_{i,\text{joint}}(m, p_i, w_i)$, becomes
\begin{align}\label{eqn:p_1_joint}
    p_{i,\text{joint}}(m,p_i,w_i) =& w_i\sum_{k=k_i^*}^m \binom{m}{k}p_i^k(1-p_i)^{m-k} \\ &+ (1-w_i)\!\!\!\!\!\sum_{k=m-k_i^*+1}^{m}\!\!\binom{m}{k} p_i^{k} (1-p_i)^{m-k}. \nonumber
\end{align}
As $m$ gets larger in (\ref{eqn:p_1_joint}), by using the Central Limit Theorem (with Gaussian approximation with continuity correction as in \cite[page 120]{hajek2020probability}), we can approximate $ p_{i,\text{joint}}(m,p_i,w_i)$ as  
\begin{align}\label{eqn:p_1_joint_apprx}
    \tilde{p}_{i,\text{joint}}(m,p_i,w_i) =& w_i Q\left(\frac{k_i^*-m p_i -0.5}{\sqrt{mp_i(1-p_i)}}\right) \\ &+ (1-w_i)Q\left(\frac{m(1-p_i)-k_i^* +0.5}{\sqrt{mp_i(1-p_i)}}\right), \nonumber
\end{align}
where $Q(\cdot)$ is the Q-function \cite{yates2014probability}. 

Next, we introduce our response timeliness metric. 
\subsection{Response Timeliness}

In this subsection, we evaluate the timeliness of responses by measuring the inter-departure times of accurate responses from the task processor. This is equivalent to the average system time of an accurate response delivered to users. From user~$i$'s perspective, the system time begins immediately after the delivery of an accurate response to user~$i$. It includes the waiting times for query arrivals as well as the response times for the queries of all users, up to the point when the next accurate response is returned to user~$i$. We denote the random variable representing the system time of an accurate response to user $i$ as $S_i$ and our goal is to characterize its average, that is, $\mathbb{E}[S_i]$.

To characterize $S_i$, let us first find the average response time of the task processor when a query from user $i$ is admitted. We denote the random variables representing the transmission time of a query from task processor to $LLM_{i,j}$ as $T_{i,j}$ where $T_{i,j}$ has an exponential distribution with rate $\mu_{i}$. Upon receiving a query from the task processor, each $LLM_{i,j}$ has a fixed response time $t_i$. Thus, the total response time of each $LLM_{i,j}$ has a shifted exponential distribution given by $T_{i,j}+t_i$. The overall response time of the task processor to user $i$'s query is given by $T_i  = t_i+ \max_{j\in\{1,\cdots,m\}} \{T_{i,j}\}$. Then, the expected value of $T_i$ is given by $\mathbb{E}[T_i]  \!=\! t_i\!+\! \mathbb{E}[\max_{j\in\!\{1,\cdots,m\}} \!\{T_{i,j}\}]$ which is equal to 
\begin{align}\label{eqn:service_time_user_i}
    \mathbb{E}[T_i]  = t_i+ \frac{1}{\mu_i} \sum_{j=1}^{m}\frac{1}{j}.
\end{align}
For large values of $m$, by using the harmonic series sum, $\mathbb{E}[T_i]$ in (\ref{eqn:service_time_user_i}) can be approximated as $\mathbb{E}[T_i] \approx t_i+ \frac{\log{m} +\gamma}{\mu_i} $ where $\gamma = 0.577$ \cite{weisstein2002euler}.
Then the average response time of the task processor to all users is given by 
\begin{align}
    \mathbb{E}[T] =\sum_{i=1}^{n} \frac{\lambda_i }{\sum_{j=1}^{n}\lambda_j}\mathbb{E}[T_i].  
\end{align}
After serving a user, the task processor waits for the next query arrival. Let
$W_i$ denote the waiting time for user $i$'s query, where we have $W_i\sim exp(\lambda_i)$. Then, the overall waiting time for the next query from any user, denoted by $W$, is given by $W = \min_{i=1,\cdots, n}\{W_i\}$, where $W$ has an exponential distribution with rate $\sum_{i=1}^n \lambda_i$. At the end of $W$, the next query belongs to user $i$ with probability $P(W=W_i) = \frac{\lambda_i}{\sum_{j=1}^n \lambda_j}$. If the incoming query, when the processor is idle, belongs to a user other than user $i$, the processor starts serving that user. 
Let $T_{-i}$ denote this busy time, i.e., the response time of the task processor to users other than user $i$. Then we have 
\begin{align}
 \mathbb{E}[T_{-i}] = \sum_{\ell=1, \ell\neq i}^{n}   \frac{\lambda_\ell  }{\sum_{j=1, j\neq i}^{n}\lambda_j} \mathbb{E}[T_\ell].
\end{align}
After serving that other user, the task processor becomes idle again and starts waiting for the next query arrival which can be either from user $i$ with probability $\frac{\lambda_i}{\sum_{j=1}^{n}\lambda_j}$ or from others with probability $\frac{\sum_{j=1, j\neq i}^{n}\lambda_j}{\sum_{j=1}^{n}\lambda_j}$. Thus, there is a geometric random variable $Y_i$ with success probability $\frac{\lambda_i}{\sum_{j=1}^{n}\lambda_j} $. The total waiting time for user $i$'s next query arrival to the idle task processor becomes $\bar{W}_{i} = \sum_{j=1}^{Y_i-1} T_{-i}(j) + \sum_{j=1}^{Y_i} W(j)$ where we have 
\begin{align}\label{eqn:total_waiting}
\mathbb{E}[\bar{W}_{i}] = \frac{1}{\lambda_i}\left(\sum_{\ell=1, \ell\neq i}^{n}\lambda_\ell \left(t_\ell+ \frac{1}{\mu_\ell} \sum_{j=1}^{m}\frac{1}{j}\right) +1\right)
 \end{align}

Next, we focus our attention to find the closed form expression for the system time of user $i$, $S_i$. User~$i$'s query enters the task processor, which requires $T_i$ units of time to process. If the response is correct, the system time $S_i$ equals $\bar{W}_{i} + T_i$. However, if the response is inaccurate, the user must wait another $\bar{W}_i$ for the next query to be admitted to the processor, followed by an additional processing time $T_i$. This process repeats itself until the correct response to user $i$'s query is obtained. This corresponds to another geometric random variable $X_i$ with success probability $p_{i,\text{joint}}(m, p_i, w_i)$, representing the number of attempts until a correct response is obtained. Thus, the total system time $S_i$ is given by
  $S_i =\sum_{j=1}^{X_i} T_i(j) + \sum_{j=1}^{X_i} \bar{W}_i(j). $
As a result, we have
\begin{align}\label{eqn:total_service_time}
    \mathbb{E}[S_i] = \frac{\sum_{\ell=1}^{n}\lambda_\ell \left(t_\ell+ \frac{1}{\mu_\ell} \sum_{j=1}^{m}\frac{1}{j}\right)+1}{\lambda_i p_{i,\text{joint}}(m,p_i,w_i)}.
\end{align}
The task processor aims to optimize timeliness, i.e., the expected system time over all users, denoted by $\mathbb{E}[S]$. Since an incoming query belongs to user $i$ with probability $\frac{\lambda_i}{\sum_{\ell=1}^{n} \lambda_\ell}$, we have $\mathbb{E}[S] =\sum_{i=1}^{n} \frac{\lambda_i}{\sum_{\ell=1}^{n}\lambda_\ell }\mathbb{E}[S_i] $ which is given by
\begin{align}\label{eqn:total_service_time_all_users}
    \!\!\mathbb{E}[S] \!\!=\!\!\!\sum_{i=1}^{n}\!\!\frac{1}{p_{i,\text{joint}}(m,p_i,w_i)}\!\!\!\left(\!\!\frac{\sum_{\ell=1}^{n}\!\!\lambda_\ell\! \left(t_\ell\!\!+\!\! \frac{1}{\mu_\ell} \!\sum_{j=1}^{m}\!\!\frac{1}{j}\right)\!+\!1}{\sum_{\ell=1}^{n}\lambda_\ell}\!\!\right)\!\!.\!\!
\end{align}
Next, we formulate the optimization problem to balance information accuracy and response timeliness in this system. 
\subsection{Problem Formulation}
In this work, we aim to deliver timely and accurate responses to users by optimizing a weighted average of response accuracy and system time through the selection of $m$. For this, we formulate the following optimization problem: \begin{align}\label{Eqn:Obj_fnc}
    \min_{m \in {\mathbb{Z}^+}} \sum_{i=1}^{n} \frac{1}{p_{i,\text{joint}}(m,p_i,w_i)} + \theta \mathbb{E}[S].
\end{align}  
Here, $\theta\geq 0$ is the weight parameter assigned to the system time of the users, $m$ is a positive integer, and $\mathbb{E}[S]$ is the average system time of this system given in (\ref{eqn:total_service_time_all_users}). 

In the next section, we present our solution approach for determining a value of $m$ that yields a sub-optimal solution to the problem in (\ref{Eqn:Obj_fnc}).

\section{Optimization of the Number of LLMs}\label{Sec:Opt_of_LLMs}
In this section, we provide our optimization strategy for the number of LLMs in each cluster, $m$, to solve the problem given in (\ref{Eqn:Obj_fnc}). We develop our solution method under the assumption that $p_i> 0.5$ for all $i$.\footnote{The solution can be easily extended to handle cases where $p_i < 0.5$ as well. Note that when $p_i < 0.5$, the accuracy of the aggregated response converges to 0. Thus, by taking the opposite decision of the aggregated response, we can obtain an accurate response. Therefore, the solutions obtained in this method are symmetric around $p_i = 0.5$ where having an accuracy close to 1 or 0 helps to get higher aggregated response accuracy whereas having accuracy close to $p_i = 0.5$ performs the worst.} First, we explicitly write the optimization problem in (\ref{Eqn:Obj_fnc}) as  
\begin{align}\label{Eqn:Obj_fnc_exp}
    \!\!\sum_{i=1}^{n} \!\!\frac{1}{p_{i,\text{joint}}(m,\!p_i,\!w_i)}\!\!\left(\!\!1 \! \!+ \!\theta \frac{\sum_{\ell=1}^{n}\!\!\lambda_\ell\! \!\left(t_\ell\!+\! \frac{1}{\mu_\ell} \sum_{j=1}^{m}\!\frac{1}{j}\right)\!+\!1}{\sum_{\ell=1}^{n}\!\!\lambda_\ell} \!\!\right)\!\!.\! \!\!
\end{align}
Then, for large values of $m$, by using the harmonic series sum and the fact that $p_{i,\text{joint}}(m,p_i,w_i)\approx \tilde{p}_{i,\text{joint}}(m,p_i,w_i)$ (as in (\ref{eqn:p_1_joint_apprx})), we can approximate (\ref{Eqn:Obj_fnc_exp}) as
\begin{align*}
    \!\!\sum_{i=1}^{n} \!\frac{1}{\tilde{p}_{i,\text{joint}}(m,p_i,w_i)}\!\left(\!1 \! + \!\theta \frac{\sum_{\ell=1}^{n}\!\lambda_\ell\! \left(t_\ell\!+\! \frac{\log (m) +\gamma }{\mu_\ell} \right)\!+\!1}{\sum_{\ell=1}^{n}\lambda_\ell} \!\right)\!. \!\!
\end{align*}
After relaxing $m$ and allowing it to take nonnegative real values, i.e., $m\in \mathbb{R}^+$ and $m\geq 1$, the problem becomes 
\begin{align}\label{Eqn:Opt_prob_2}
       \min_{m\geq 1} \sum_{i=1}^{n} \frac{1}{\tilde{p}_{i,\text{joint}}(m,p_i,w_i)}\!\!\left(\!1 \! + \!\theta \frac{\sum_{\ell=1}^{n}\!\!\lambda_\ell\! \left(t_\ell\!\!+\!\! \frac{\log (m) +\gamma }{\mu_\ell} \right)\!\!+\!\!1}{\sum_{\ell=1}^{n}\lambda_\ell} \!\right). \!\!
\end{align}
Before moving to the solution of the problem, let us first characterize the behavior of the objective function in (\ref{Eqn:Opt_prob_2}) with respect to various system parameters. For that, in the following lemma, we show that for sufficiently large $m$, $\tilde{p}_{i,\text{joint}}(m,p_i,w_i)$ is concave with respect to $m$.  

\begin{lemma}\label{Lemma_1}
    For $p_i >0.5  $, if we have 
    \begin{align}\label{eqn:concavity_cond}
        m\geq \frac{1}{p_i-0.5} \left|\frac{\log\left(\frac{1-w_i}{w_i}\right)}{2\log\left(\frac{1-p_i}{p_i}\right)} -0.5\right|, 
    \end{align}
then, $\tilde{p}_{i,\text{joint}}(m,p_i,w_i)$ is a concave function of $m$. Otherwise, it is neither a concave nor a convex function of $m$.    
\end{lemma}
\begin{Proof}
We begin our proof by inserting $k_i^*$ given in (\ref{eqn:optimum_k_i}) into $\tilde{p}_{i,\text{joint}}(m,p_i,w_i)$ given in (\ref{eqn:p_1_joint_apprx}), which yields
\begin{align}\label{eqn:p_1_joint_apprx_v2}
    \tilde{p}_{i,\text{joint}}&(m,p_i,w_i) \!\!= w_i Q\left(B_i\left(\sqrt{m}(0.5-p_i)+\frac{A_i}{\sqrt{m}}\right)\right) \nonumber\\&~~~~+ (1\!-\!w_i)Q\left(B_i\left(\sqrt{m}(0.5\!-\!p_i)\!-\!\frac{A_i}{\sqrt{m}}\right)\right)\!,\!\!
\end{align}
where $A_i =\frac{\log\left(\frac{1-w_i}{w_i}\right)}{2\log\left(\frac{1-p_i}{p_i}\right)} -0.5$ and $B_i = 1/(\sqrt{mp_i(1-p_i)})$. Since $Q(x)$ is concave when $x<0$ and convex when $x>0$, $\tilde{p}_{i,\text{joint}}(m,p_i,w_i)$ is concave when $m\geq \frac{|A_i|}{p_i-0.5}$. When this condition is not satisfied, since $m(0.5-p_i)<0$, the argument in one of the $Q$-functions becomes negative while that of the other one remains positive. Thus, when the condition in (\ref{eqn:concavity_cond}) is not satisfied, $\tilde{p}_{i,\text{joint}}(m,p_i,w_i)$ is neither convex nor concave in $m$, which completes the proof.  
\end{Proof}

As a result of Lemma~\ref{Lemma_1}, the response accuracy term in (\ref{Eqn:Opt_prob_2}), that is, $\sum_{i=1}^{n} \frac{1}{\tilde{p}_{i,\text{joint}}(m,p_i,w_i)}$ is neither convex nor concave for small values of $m$. Further, since  $\frac{A_i}{\sqrt{m}}$ becomes negligible as $m$ increases, we can approximate $\tilde{p}_{i,\text{joint}}(m,p_i,w_i)$ in (\ref{eqn:p_1_joint_apprx}) with a single Q-function, such that, $\tilde{p}_{i,\text{joint}}(m,p_i,w_i)\approx Q\!\left(\!\frac{\sqrt{m}(0.5-p_i)}{\sqrt{p_i(1-p_i)}}\!\right)$.  Then, as $m$ increases, $\sum_{i=1}^{n} \frac{1}{\tilde{p}_{i,\text{joint}}(m,p_i,w_i)}$ decreases and becomes a convex function which saturates to $n$ eventually since $\tilde{p}_{i,\text{joint}}(m,p_i,w_i)\leq 1$.

Next, let us focus on the system time part of (\ref{Eqn:Opt_prob_2}) given by $\sum_{i=1}^{n}\frac{1}{\tilde{p}_{i,\text{joint}}(m,p_i,w_i)}\left(\frac{\sum_{\ell=1}^{n}\lambda_\ell \left(t_\ell+ \frac{\log (m) +\gamma }{\mu_\ell} \right)+1}{\sum_{\ell=1}^{n}\lambda_\ell}\right)\!$. As $m$ gets large, as we argued earlier $\sum_{i=1}^{n}\frac{1}{\tilde{p}_{i,\text{joint}}(m,p_i,w_i)}$ saturates to $n$, so that the system time is mainly dominated by the second term which increases logarithmically in $m$. In other words, for sufficiently large values of $m$, $\mathbb{E}[S]$ is a concave increasing function of $m$. However, for relatively small values of $m$, the first term $N_1(m) =\left(\sum_{\ell=1}^{n}\lambda_\ell \left(t_\ell+ \frac{\log (m) +\gamma }{\mu_\ell}\right)+1\right)/ \sum_{\ell=1}^{n}\lambda_\ell$ increases with $m$ whereas the second term $N_2(m)= \sum_{i=1}^{n}\frac{1}{\tilde{p}_{i,\text{joint}}(m,p_i,w_i)}$ decreases with $m$. To determine the behavior of the average value of the system time for small $m$, we need to check the sign of  $\frac{\partial\mathbb{E}[S]}{\partial m}$ which is given by 
\begin{align}
  \frac{\partial\mathbb{E}[S]}{\partial m}= N_1'(m)N_2(m)-N_1(m)N_2'(m), 
\end{align}
where $N_1'(m) = \frac{\partial N_1(m)}{\partial m}$ and $N_2'(m) =\frac{\partial N_2(m)}{\partial m}$. One can easily verify that depending on the system parameters such as $\lambda_i$, $t_i$, $\mu_i$, $p_i$, and $w_i$, $\mathbb{E}[S]$ can be an increasing or a decreasing function of $m$ for small values of $m$.  

Combining both parts, for any value of $\theta >0$, there exists a sufficiently large $m$ beyond which the objective function in (\ref{Eqn:Opt_prob_2}) is an increasing function of $m$. However, particularly for small values of $\theta$, the objective function in (\ref{Eqn:Opt_prob_2}) may initially decrease—as driven by the $\sum_{i=1}^{n}\frac{1}{\tilde{p}_{i,\text{joint}}(m,p_i,w_i)}$ term—and then increase with respect to $m$, potentially reaching its global minimum at a critical value of $m$. Since the problem is non-convex, in general, depending on $\theta$ and the behavior of $\mathbb{E}[S]$ and $\tilde{p}_{i,\text{joint}}(m,p_i,w_i)$,  there may exist (a single or multiple) local minima of (\ref{Eqn:Opt_prob_2}). To identify these critical values of $m$, one can implement a gradient descent algorithm with multiple initializations to increase the likelihood of converging to the global minimum.     

Next, we present comprehensive experiments to evaluate $p_{i,\text{joint}}(m,p_i,w_i)$ in practice using various pre-trained LLMs, and to illustrate the minimization of $\sum_{i\!=\!1}^{n}\! \frac{1}{p_{i,\text{joint}}(m,p_i,w_i)} \!+\! \theta \mathbb{E}[S]$ in (\ref{Eqn:Obj_fnc}) described in this section.

\section{Numerical Experiments}\label{Sec:num_result}
In this section, we present numerical experiments to validate our framework. First, we describe our experimental setup.
\subsection{Experimental Setup}
 We use fully open source seven different pre-trained LLMs: Mistral-7B-Instruct-v0.3 \cite{jiang2023mistral7b}, Llama 3.1-8B \cite{grattafiori2024llama3herdmodels}, gemma-3-4b-it \cite{gemmateam2025gemma3technicalreport}, Qwen3-8B \cite{yang2025qwen3technicalreport}, Phi-4-mini-instruct \cite{abdin2024phi4technicalreport}, Flan-T5-XL \cite{chung2022scalinginstructionfinetunedlanguagemodels}, and Falcon3-7B-Instruct\cite{almazrouei2023falconseriesopenlanguage}, with temperature set to 0.1 for all models. Thus, in our experiments, $m$ varies from 1 to 7. That is, we can have one from each of these models in each LLM cluster. The experiments are performed on the Google Colab environment with an NVIDIA A100 40GB GPU to handle the computational requirements of the models.

We evaluate models on binary verification tasks derived from several popular question answering benchmarks, including TriviaQA \cite{joshi2017triviaqalargescaledistantly}, Arc-Easy \cite{clark2018thinksolvedquestionanswering}, Arc-Challenge \cite{clark2018thinksolvedquestionanswering}, and CommonsenseQA
\cite{talmor2019commonsenseqaquestionansweringchallenge}. From each of these four datasets, we randomly select 250 samples. We then convert each sample into two distinct queries using the following procedure: 
\paragraph*{1. Positive (Correct Answer) Query} To create the positive query, we combine the original question and its corresponding ground truth answer into a single, closed-ended, i.e., yes/no, prompt. That is, we format the query as follows:
\begin{quote}
\textit{``Given the provided context, for the question `[Original Question]', is the answer `[Correct Answer]'?''}
\end{quote}
This newly generated prompt, which presents a factual statement, then has a ground truth label of \textit{`true'.} The LLM's task is essentially to confirm this statement against the context.
\begin{figure}[t]
\centering
\includegraphics[width=0.41\textwidth]{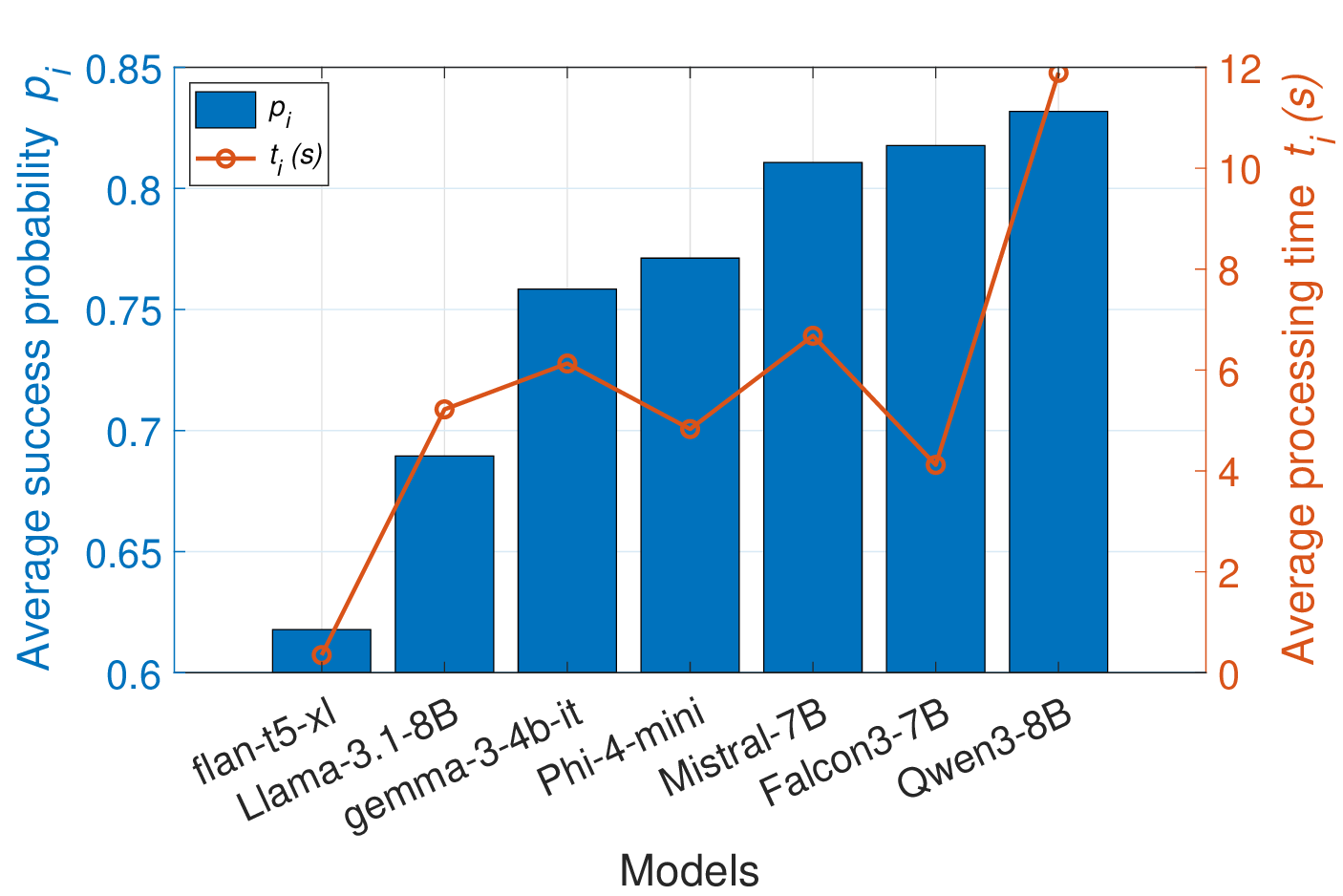}
\vspace{-0.35cm}
\caption{\!\!Average processing time vs average success probability for each model.}
\label{fig:time_vs_accuracy}
\vspace{-0.5cm}
\end{figure}
\paragraph*{2. Negative (Incorrect Answer) Query} To create the corresponding negative query, we use the same original question and context, but this time we pair them with an incorrect answer deliberately. Here, the incorrect answer that we choose is the correct answer of another randomly selected question from the selected samples. As a result, we create a new query using the same format:
\begin{quote}
\textit{``Given the provided context, for the question `[Original Question]', is the answer `[Incorrect Answer]'?"}
\end{quote}
This query, which presents a false statement, has a ground truth label of \textit{`false'.} The LLM's task is to correctly identify the discrepancy and disagree.
\begin{figure*}[!t]
\centering
\includegraphics[width=0.95\textwidth]{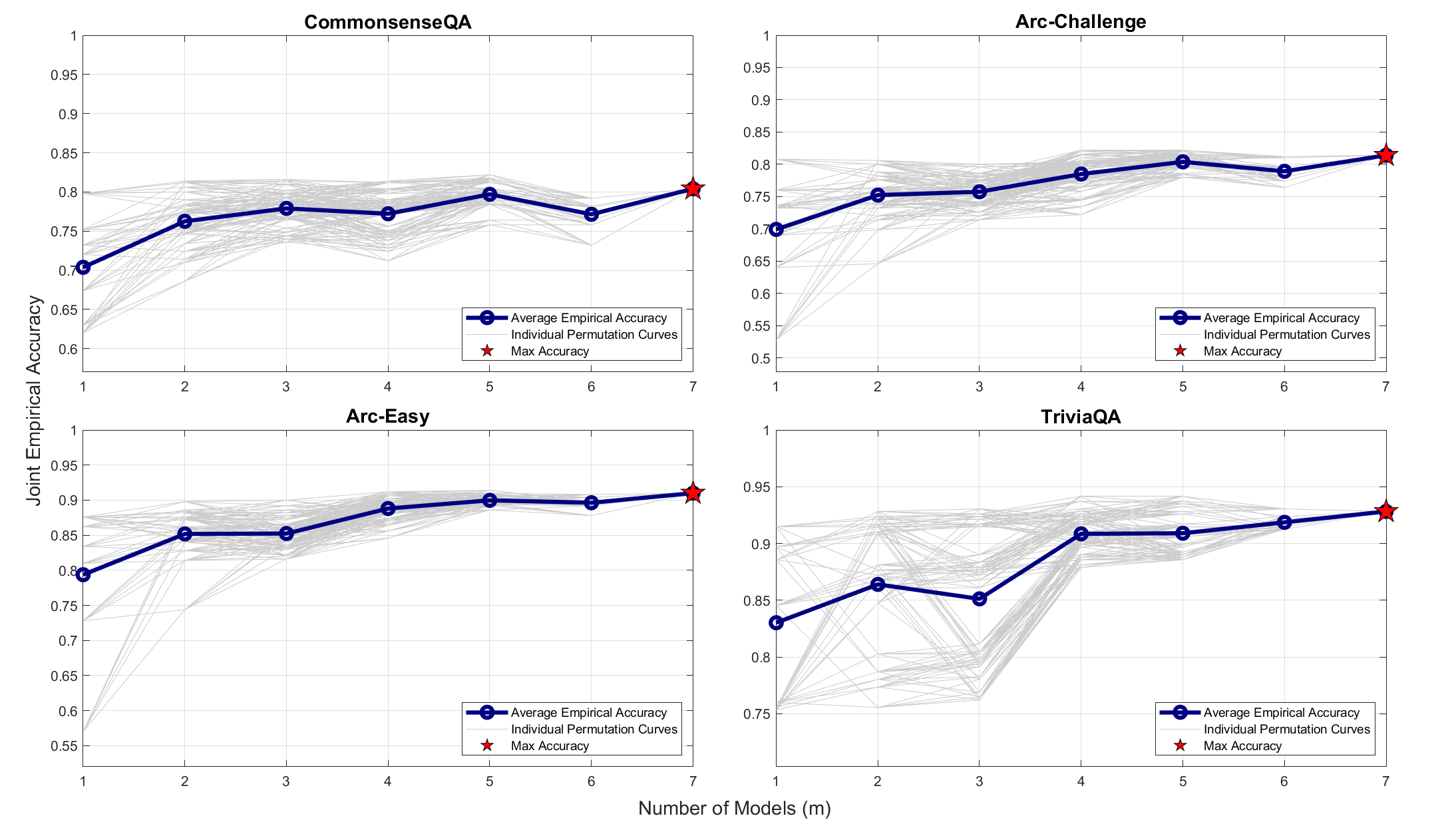}
\vspace{-0.35cm}
\caption{Performance on various QA datasets. Our joint decision making framework can utilize the collective capabilities of multiple agents to produce more reliable outcomes than any single LLM.}
\label{fig:Acc_plot}
\vspace{-0.5cm}
\end{figure*}

In our first experiment, we evaluate the individual accuracy of each of the seven models. A key aspect of this evaluation is making sure that a model's intended binary output is classified as `true' or `false'. Instead of relying on exhaustive human annotation, we employ a two-way answer extraction pipeline designed to process model outputs consistently. For models capable of generating direct binary responses, e.g., `yes' or `no', we use a simple rule-based parser. This parser scans the first 20 tokens of the response for binary expressions, such as `true' or `false', `yes' or `no', and determines the final binary output accordingly. For models that produce more verbose, conversational responses, we employ an extractor LLM (Qwen3-8B) to analyze the entire output and classify the intended answer as either `true' or `false' \cite{zheng2023judgingllmasajudgemtbenchchatbot, chiang2023largelanguagemodelsalternative}. This automated process enables us to establish a baseline individual accuracy, $p_i$, for each model. Additionally, to assess the time overhead of the individual models, we empirically analyze their processing times for each benchmark.

Finally, we evaluate the joint performance of multi-LLM ensembles to measure the empirical accuracy of our system. To provide an unbiased assessment and reduce any order-dependent effects from a fixed model sequence, our analysis considers all $m!$ possible permutations of the $m$ models within a cluster, where $m$ is at most seven in our setup. For each permutation, we compute the joint empirical accuracy for incrementally growing ensemble sizes, from $m = 1$ to $7$. As described above, for each sample, we generate pair of queries: one `positive' (the correct statement) and one `negative' (an incorrect statement). This experimental design results in a balanced dataset where the prior probability of a query being true is equal to probability of it being false, i.e., $w_i = 0.5$. Consequently, the collective decision for each ensemble is based on a simple majority vote, which is a special case of our optimal MAP estimator given in (\ref{eqn:optimum_k_i}) when $w_i = 0.5$, setting the decision threshold $k^*$ to $m/2$. The resulting accuracy curves from all permutations are then averaged to produce a single ``average empirical accuracy" curve shown in Fig.~\ref{fig:Acc_plot}.

\begin{table*}[!t]
\centering
\caption{Model Performance on Various Benchmarks}
\label{tab:individual_performance}\vspace{-0.2cm}
\begin{tabular}{l c c c c c}
\toprule
\textbf{Model} & \textbf{Arc-Challenge} & \textbf{Arc-Easy} & \textbf{CommonsenseQA} & \textbf{TriviaQA} & \textbf{ Processing Time $t_i$ (s)}\\
\midrule
Mistral-7B-Instruct-v0.3 & 73.2\% & 86.2\% & 75.2\% & 89.7\% & 6.68 \\
Llama 3.1-8B & 64.0\% & 72.8\% & 63.0\% & 76.0\% & 5.22\\
gemma-3-4b-it & 73.6\% & 81.0\% & 73.2\% & 75.6\% & 6.13 \\

Qwen3-8B & 80.8\% & 87.6\% & 79.8\% & 84.5\% & 11.89\\
Phi-4-mini-instruct & 69.0\% & 83.4\% & 67.4\% & 88.6\% & 4.83\\
Flan-T5-XL & 52.8\% & 57.0\% & 62.0\% & 75.3\% & 0.35\\
Falcon3-7B-Instruct & 76.0\% & 87.6\% & 72.0\% & 91.5\% & 4.12\\
\midrule
7-LLM Ensemble& \textbf{81.4}\% & \textbf{91.0}\% & \textbf{80.4}\% & \textbf{92.8}\% & -\\
\bottomrule
\end{tabular}
\vspace{-0.5cm}
\end{table*}

\subsection{Performance Evaluation}
In this section, we evaluate our framework by analyzing the joint accuracy of multi-LLM ensembles and the trade-off between response accuracy and timeliness. The comprehensive results of our experiments are presented in Table~\ref{tab:individual_performance} with visualizations shown in Figs.~\ref{fig:time_vs_accuracy} and \ref{fig:Acc_plot}. 

We first empirically evaluate the average performance of each of the seven models across all benchmarks. Fig.~\ref{fig:time_vs_accuracy} shows the average success probabilities $p_i$ plotted with the corresponding average processing times $t_i$. While our analytical framework assumes identical success probabilities $p_i$ and processing times $t_i$ for all LLMs within a given cluster, empirical results, as expected, indicate notable differences in both accuracy and processing times among the models. However, as we will show in Fig.~\ref{fig:commonsenseQA}, our theoretical results still provide effective lower and upper bounds when applied to these LLMs.  Models such as Flan-T5-XL offer the fastest responses (lowest observed $t_i$), but at the expense of reduced accuracy (lowest observed $p_i$), whereas Qwen3-8B achieves the highest accuracy with the longest processing time. For real-time applications, models with low processing time may be preferable, while tasks prioritizing accuracy can justify the use of slower, more accurate models like Qwen3-8B or Mistral-7B. Some models, such as Falcon3-7B, strike a balance between accuracy and processing time, making them strong general-purpose candidates. Depending on the desired system behavior, individual LLMs can be selected accordingly when forming an LLM cluster.

Our primary finding, illustrated across the four subplots in Fig.~\ref{fig:Acc_plot}, is that aggregating responses from multiple models yields improved performance. For all QA benchmarks, the average empirical accuracy (the dark blue line) demonstrates a clear upward trend as the number of models in the ensemble $m$ increases. This validates the core presumption that our joint decision making framework can utilize the collective capabilities of multiple agents to produce more reliable outcomes than any single LLM. The light gray lines, representing every possible model permutation, i.e., a subset of models for a given $m$, highlight that while the performance of small ensembles can be sensitive to the specific models chosen, the accuracy becomes more stable and robust as the ensemble grows.  

Furthermore, we observe that the composition of the ensemble is a critical factor. Ensembles composed of models with comparable accuracy levels tend to yield more significant gains in joint accuracy. Conversely, when a single high performing \textit{expert} model is paired with multiple lower accuracy agents, the weaker models sometimes act as noise, limiting the ensemble's potential by pulling the collective decision away from the expert's correct prediction. Table~\ref{tab:individual_performance} validates these improvements. On the Arc-Challenge benchmark, for instance, the best individual model (Qwen3-8B) achieves an accuracy of 80.8\%, whereas the 7-LLM ensemble reaches a slightly higher accuracy of 81.4\%. On the other hand, the benefit of joint response generation is more notable on the Arc-Easy benchmark, where the ensemble's accuracy of 91.0\% surpasses the top performing single model (Falcon3-7B-Instruct and Qwen3-8B) by 3.4 points. In all evaluated cases, the maximum performance, marked by the red star in Fig.~\ref{fig:Acc_plot} is achieved with the full 7-model ensemble.

\subsection{Comparison of $p_{i,\text{joint}}(m,p_i,w_i)$ with Empirical Results}
\begin{figure}[t]
\centering
\includegraphics[width=0.40\textwidth]{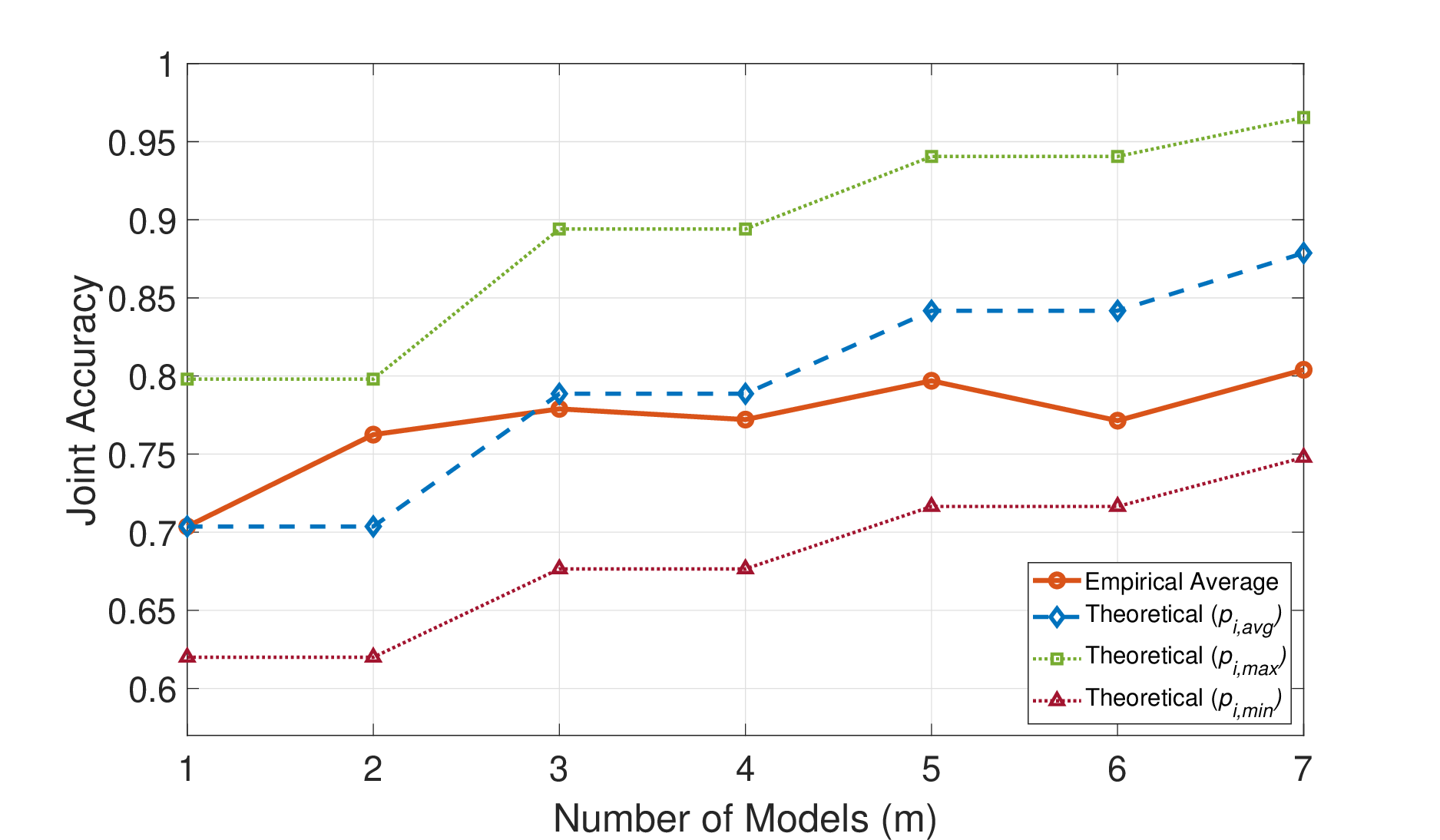}
\vspace{-0.4cm}
\caption{Comparison of empirical and theoretical joint accuracy performance, i.e., success probability, on the CommonsenseQA dataset.}
\label{fig:commonsenseQA}
\vspace{-0.5cm}
\end{figure}

In this experiment, we compare the theoretical information accuracy $p_{i,\text{joint}}(m,p_i,w_i)$ that we derived in (\ref{eqn:p_1_joint}) with the empirical performance obtained for the CommonsenseQA dataset. We obtain our theoretical results on $p_{i,\text{joint}}(m,p_i,w_i)$ under two main assumptions: 1) responses collected from the LLMs are independent from each other, and 2) LLMs within a cluster have the same accuracy $p_i$. However, as we observe in previous experiments, the individual accuracy of LLMs may differ from each other. For example, in Table~\ref{tab:individual_performance}, for the CommonsenseQA dataset, individual accuracy of the LLMs varies from $p_6 = 0.62$ (for Flan-T5-XL) to $p_4 = 0.798$ (for Qwen3-8B). Moreover, due to the lack of transparency regarding the internal architecture and training processes of the LLMs, it is unclear whether their responses can be assumed to be statistically independent. However, to derive a lower bound on information accuracy, we begin by computing $p_{i,\text{joint}}(m,p_i,w_i)$, using the lowest-accuracy model, $p_6 = 0.62$ (for Flan-T5-XL), as indicated by the plot with the triangular markers in Fig.~\ref{fig:commonsenseQA}. Similarly, to provide an upper bound we compute $p_{i,\text{joint}}(m,p_i,w_i)$ with the highest-accuracy model, namely, $p_4 = 0.798$ (for Qwen3-8B) as indicated by the plot with the square markers in Fig.~\ref{fig:commonsenseQA}. These two curves provide upper and lower bounds on the information accuracy of the empirical results. If the difference between the lowest and the highest accuracy model is low, these bounds become tighter. During our experiments, we observe that if we take the mean of the individual accuracy of the LLMs, that is, $p_{i,\text{avg}} = \frac{1}{m}\sum_{j=1}^{m} p_{i,j}$ (where $p_{i,j}$ is the accuracy of LLM $j$ to user $i$'s query) and compute $p_{i,\text{joint}}(m,p_i,w_i)$ in (\ref{eqn:p_1_joint}) by using $p_{i,\text{avg}}$, the theoretical results are much closer to the empirical accuracy as shown by the diamond markers in Fig.~\ref{fig:commonsenseQA}.

\subsection{Optimization of $m$}

\begin{figure}[t]
    \begin{center}
        \subfigure[]{%
        \includegraphics[width=0.49\columnwidth]{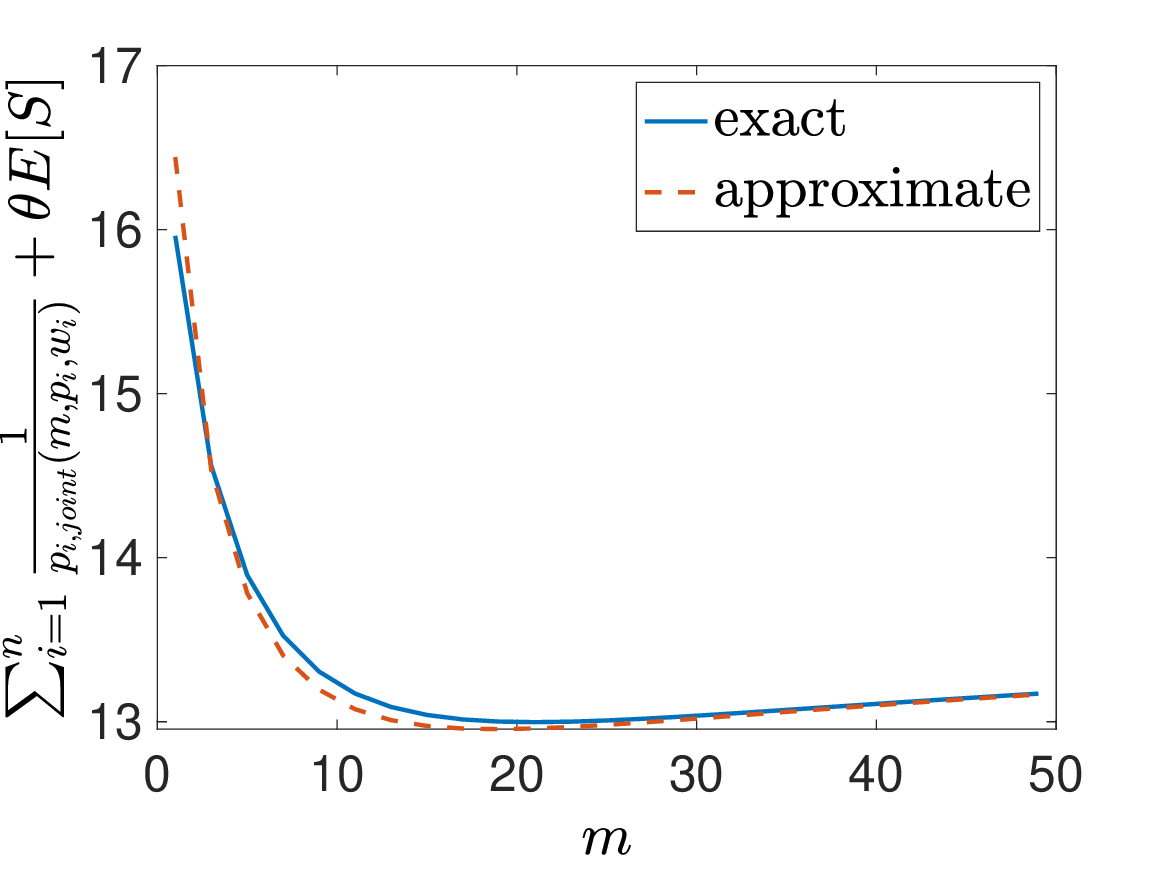}}
        \subfigure[]{%
        \includegraphics[width=0.49\columnwidth]{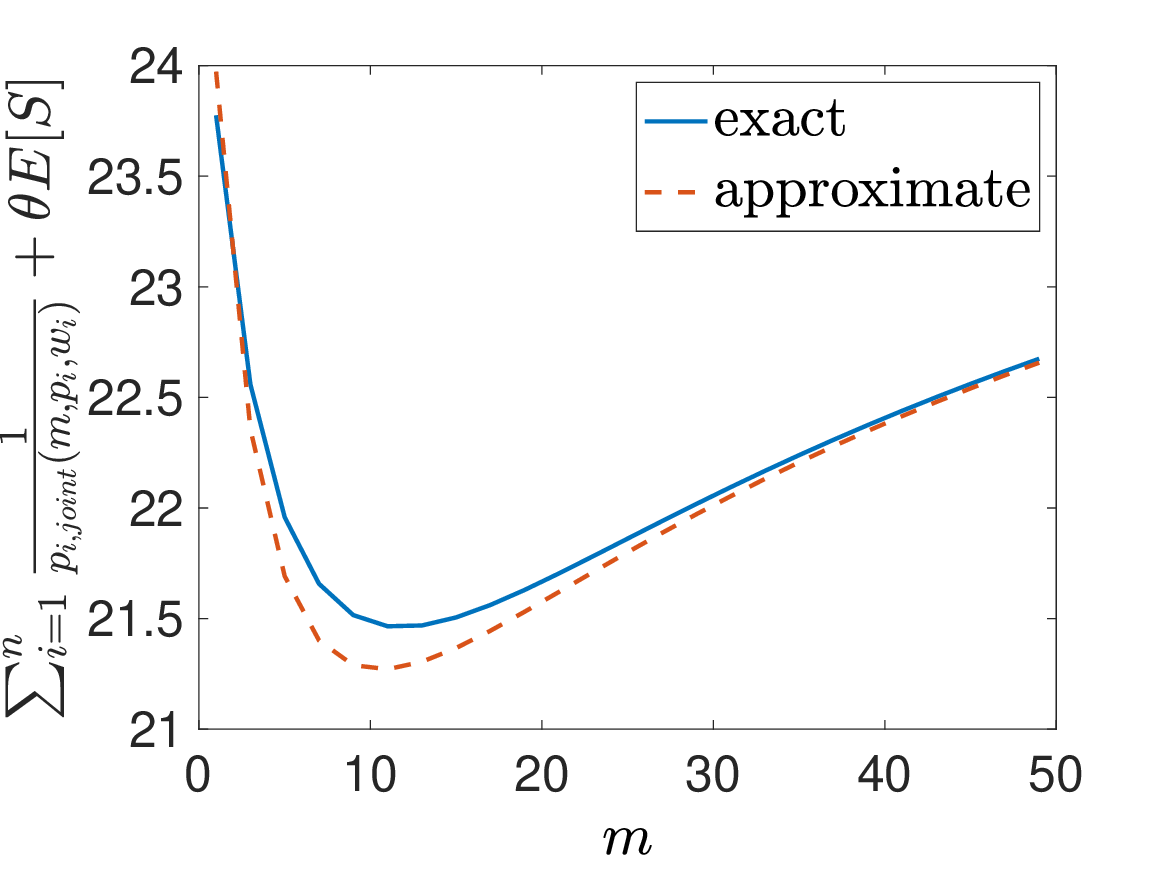}}
    \end{center}
    \vspace{-0.35cm}
    \caption{The objective function in (\ref{Eqn:Obj_fnc}) and its approximate in (\ref{Eqn:Opt_prob_2}) with respect to $m$ when (a) $\theta = 0.1$ and (b) $\theta = 0.4$. }
    \label{fig:sim_1}
     \vspace{-0.5cm}
\end{figure}
In our next simulation result, we consider the optimization of the objective function given in (\ref{Eqn:Obj_fnc}) and its approximation in (\ref{Eqn:Opt_prob_2}) through $m$. With this goal, we choose $n=10$, $w_i = 0.5$, $p_i = 0.7 + (i-1)/90$, $t_i = 1 + (i-1)/9$, and $\mu_i = 2 + 2(i-1)/9 $ for all $i \in [1,n]$. We evaluate the objective function and its approximation for two different values of $\theta = \{0.1, 0.4\}$ for $m =1,\dots, 50$ and plot them in Fig.~\ref{fig:sim_1}. In both Figs.~\ref{fig:sim_1}(a) and (b), we observe that the approximate objective function derived in (\ref{Eqn:Opt_prob_2}) closely follows the exact objective function in (\ref{Eqn:Obj_fnc}). When $\theta = 0.1$, we prioritize the information accuracy compared to the average system time. Since the information accuracy increases with $m$, we see in Fig.~\ref{fig:sim_1} (a) that the minimum value of the objective function is attained when $m^*=21$ while its approximation in (\ref{Eqn:Opt_prob_2}) attains the minimum when $m^*=19$. When $\theta = 0.4$, we give more importance to the average system time which increases with $m$. For that reason, the minimum value of the objective function and its approximation is obtained when $m^*=11$ as shown in Fig.~\ref{fig:sim_1}(b). Thus, one can choose a suitable $\theta$ value that strikes a balance between information accuracy and the average system time. We observe that, with the chosen parameter settings in this simulation, the objective function exhibits a well-behaved U-shape, allowing the global optimum of $m$ to be efficiently computed using a gradient-descent-type algorithm. However, as previously discussed, the objective function is generally non-convex, hence, it may exhibit multiple local minima.
\vspace{-0.6em}
\section{Discussion and  Conclusion}\label{sec:conc}
\vspace{-0.4em}
In this work, we studied the trade-off between the information accuracy and response timeliness in networked LLM systems. In our model, each user's query belongs to a distinct category and is forwarded to a corresponding cluster consisting of $m$ LLMs. After collecting responses from these $m$ LLMs, the task processor aggregates them into a final response. Assuming that the responses are independent and that all LLMs within the same cluster have equal accuracy, we derived a closed form expression for the joint accuracy and the timeliness of the final response. Our analysis showed that while information accuracy increases with $m$, the average system time also increases, highlighting a fundamental trade-off. To address this, we formulated an optimization problem that balances response timeliness and accuracy. One can employ a gradient-descent-type algorithm to identify a suitable value of $m$. However, due to the non-convex nature of the objective function, the algorithm may converge to local minima. To improve the likelihood of reaching the global minimum, the algorithm should be initialized from multiple starting points. 

In our extensive simulation results on various pre-trained LLMs, we indeed observed that the accuracy of the joint response surpasses that of the individual models. This improvement is more pronounced when the individual LLMs have similar accuracy levels. Conversely, when an ensemble includes an expert LLM with significantly higher accuracy than the others, the accuracy of the joint response may not substantially exceed that of the expert alone, suggesting that additional factors may influence the overall information accuracy in such cases. 

Despite being developed under simplifying assumptions, our framework offers a foundation for understanding and formalizing information accuracy and timeliness in networked LLM systems. It potentially opens up several promising research directions, including scenarios involving heterogeneous LLMs, as well as settings where the task processor lacks prior knowledge of the individual LLMs' accuracies.

\bibliographystyle{IEEEtran}
\bibliography{references}

\begin{thebibliography}{10}
\providecommand{\url}[1]{#1}
\csname url@samestyle\endcsname
\providecommand{\newblock}{\relax}
\providecommand{\bibinfo}[2]{#2}
\providecommand{\BIBentrySTDinterwordspacing}{\spaceskip=0pt\relax}
\providecommand{\BIBentryALTinterwordstretchfactor}{4}
\providecommand{\BIBentryALTinterwordspacing}{\spaceskip=\fontdimen2\font plus
\BIBentryALTinterwordstretchfactor\fontdimen3\font minus \fontdimen4\font\relax}
\providecommand{\BIBforeignlanguage}[2]{{%
\expandafter\ifx\csname l@#1\endcsname\relax
\typeout{** WARNING: IEEEtran.bst: No hyphenation pattern has been}%
\typeout{** loaded for the language `#1'. Using the pattern for}%
\typeout{** the default language instead.}%
\else
\language=\csname l@#1\endcsname
\fi
#2}}
\providecommand{\BIBdecl}{\relax}
\BIBdecl

\bibitem{brown2020language}
T.~Brown, B.~Mann, N.~Ryder, M.~Subbiah, J.~D. Kaplan, P.~Dhariwal, A.~Neelakantan, P.~Shyam, G.~Sastry, A.~Askell \emph{et~al.}, ``Language models are few-shot learners,'' \emph{Advances in neural information processing systems}, vol.~33, pp. 1877--1901, 2020.

\bibitem{touvron2023llama}
H.~Touvron, T.~Lavril, G.~Izacard, X.~Martinet, M.-A. Lachaux, T.~Lacroix, B.~Rozi{\`e}re, N.~Goyal, E.~Hambro, F.~Azhar \emph{et~al.}, ``Llama: Open and efficient foundation language models,'' \emph{arXiv preprint arXiv:2302.13971}, 2023.

\bibitem{grattafiori2024llama3herdmodels}
\BIBentryALTinterwordspacing
A.~Grattafiori, A.~Dubey, A.~Jauhri, and et~al., ``The {Llama} 3 herd of models,'' 2024. [Online]. Available: \url{https://arxiv.org/abs/2407.21783}
\BIBentrySTDinterwordspacing

\bibitem{yang2025qwen3technicalreport}
\BIBentryALTinterwordspacing
A.~Yang, A.~Li, B.~Yang, B.~Zhang, B.~Hui, and et~al., ``Qwen3 technical report,'' 2025. [Online]. Available: \url{https://arxiv.org/abs/2505.09388}
\BIBentrySTDinterwordspacing

\bibitem{luo2025toward}
H.~Luo, Y.~Liu, R.~Zhang, J.~Wang, G.~Sun, D.~Niyato, H.~Yu, Z.~Xiong, X.~Wang, and X.~Shen, ``Toward edge general intelligence with multiple-large language model {(Multi-LLM)}: Architecture, trust, and orchestration,'' \emph{arXiv preprint arXiv:2507.00672}, 2025.

\bibitem{cemri2025multiagentllmsystemsfail}
\BIBentryALTinterwordspacing
M.~Cemri, M.~Z. Pan, S.~Yang, L.~A. Agrawal, B.~Chopra, R.~Tiwari, K.~Keutzer, A.~Parameswaran, D.~Klein, K.~Ramchandran, M.~Zaharia, J.~E. Gonzalez, and I.~Stoica, ``Why do multi-agent {LLM} systems fail?'' 2025. [Online]. Available: \url{https://arxiv.org/abs/2503.13657}
\BIBentrySTDinterwordspacing

\bibitem{wang2024mixture}
J.~Wang, J.~Wang, B.~Athiwaratkun, C.~Zhang, and J.~Zou, ``Mixture-of-agents enhances large language model capabilities,'' \emph{arXiv preprint arXiv:2406.04692}, 2024.

\bibitem{li2025smoa}
D.~Li, Z.~Tan, P.~Qian, Y.~Li, K.~Chaudhary, L.~Hu, and J.~Shen, ``{SMoA: Improving} multi-agent large language models with sparse mixture-of-agents,'' in \emph{Pacific-Asia Conference on Knowledge Discovery and Data Mining}.\hskip 1em plus 0.5em minus 0.4em\relax Springer, 2025, pp. 54--65.

\bibitem{xie2025rmoaoptimizingmixtureofagentsdiversity}
\BIBentryALTinterwordspacing
Z.~Xie, C.~Han, J.~Shi, W.~Cui, X.~Zhao, X.~Wu, and J.~Zhao, ``{RMoA: Optimizing} mixture-of-agents through diversity maximization and residual compensation,'' 2025. [Online]. Available: \url{https://arxiv.org/abs/2505.24442}
\BIBentrySTDinterwordspacing

\bibitem{du2023improving}
Y.~Du, S.~Li, A.~Torralba, J.~B. Tenenbaum, and I.~Mordatch, ``Improving factuality and reasoning in language models through multiagent debate,'' \emph{arXiv preprint arXiv:2305.14325}, 2023.

\bibitem{liang2023encouraging}
T.~Liang, Z.~He, W.~Jiao, X.~Wang, Y.~Wang, R.~Wang, Y.~Yang, S.~Shi, and Z.~Tu, ``Encouraging divergent thinking in large language models through multi-agent debate,'' \emph{arXiv preprint arXiv:2305.19118}, 2023.

\bibitem{ye2025xmasbuildingmultiagentsystems}
\BIBentryALTinterwordspacing
R.~Ye, X.~Liu, Q.~Wu, X.~Pang, Z.~Yin, L.~Bai, and S.~Chen, ``X-mas: Towards building multi-agent systems with heterogeneous {LLMs},'' 2025. [Online]. Available: \url{https://arxiv.org/abs/2505.16997}
\BIBentrySTDinterwordspacing

\bibitem{chen2023frugalgptuselargelanguage}
\BIBentryALTinterwordspacing
L.~Chen, M.~Zaharia, and J.~Zou, ``{FrugalGPT: How} to use large language models while reducing cost and improving performance,'' 2023. [Online]. Available: \url{https://arxiv.org/abs/2305.05176}
\BIBentrySTDinterwordspacing

\bibitem{wang-etal-2025-mixllm}
\BIBentryALTinterwordspacing
X.~Wang, Y.~Liu, W.~Cheng, X.~Zhao, Z.~Chen, W.~Yu, Y.~Fu, and H.~Chen, ``{M}ix{LLM}: Dynamic routing in mixed large language models,'' in \emph{Proceedings of the 2025 Conference of the Nations of the Americas Chapter of the Association for Computational Linguistics: Human Language Technologies (Volume 1: Long Papers)}, L.~Chiruzzo, A.~Ritter, and L.~Wang, Eds.\hskip 1em plus 0.5em minus 0.4em\relax Albuquerque, New Mexico: Association for Computational Linguistics, Apr. 2025, pp. 10\,912--10\,922. [Online]. Available: \url{https://aclanthology.org/2025.naacl-long.545/}
\BIBentrySTDinterwordspacing

\bibitem{zhang2025leveraginguncertaintyestimationefficient}
\BIBentryALTinterwordspacing
T.~Zhang, A.~Mehradfar, D.~Dimitriadis, and S.~Avestimehr, ``Leveraging uncertainty estimation for efficient {LLM} routing,'' 2025. [Online]. Available: \url{https://arxiv.org/abs/2502.11021}
\BIBentrySTDinterwordspacing

\bibitem{zhang2025lightrouterefficientllmcollaboration}
\BIBentryALTinterwordspacing
Y.~Zhang, X.~Zhao, Z.~Wang, G.~Cheng, Y.~Xu, S.~Deng, and J.~Yin, ``{LightRouter: Towards} efficient {LLM} collaboration with minimal overhead,'' 2025. [Online]. Available: \url{https://arxiv.org/abs/2505.16221}
\BIBentrySTDinterwordspacing

\bibitem{srivatsa2024harnessingpowermultipleminds}
\BIBentryALTinterwordspacing
K.~A. Srivatsa, K.~K. Maurya, and E.~Kochmar, ``Harnessing the power of multiple minds: Lessons learned from {LLM} routing,'' 2024. [Online]. Available: \url{https://arxiv.org/abs/2405.00467}
\BIBentrySTDinterwordspacing

\bibitem{ding2024hybridllmcostefficientqualityaware}
\BIBentryALTinterwordspacing
D.~Ding, A.~Mallick, C.~Wang, R.~Sim, S.~Mukherjee, V.~Ruhle, L.~V.~S. Lakshmanan, and A.~H. Awadallah, ``Hybrid {LLM}: Cost-efficient and quality-aware query routing,'' 2024. [Online]. Available: \url{https://arxiv.org/abs/2404.14618}
\BIBentrySTDinterwordspacing

\bibitem{mitra2024distributed}
P.~Mitra, P.~Kaswan, and S.~Ulukus, ``Distributed mixture-of-agents for edge inference with large language models,'' \emph{arXiv preprint arXiv:2412.21200}, 2024.

\bibitem{Touri23_ISIT}
A.~Verma, A.~Sharbafchi, B.~Touri, and S.~Mohajer, ``Distributed fact checking,'' in \emph{2023 IEEE International Symposium on Information Theory (ISIT)}, 2023, pp. 2649--2654.

\bibitem{hajek2020probability}
B.~Hajek, ``Probability with engineering applications,'' \url{https://courses.grainger.illinois.edu/ece313/fa2020/probabilityJan25.pdf}, Jan. 2020, {Lecture Notes, ECE 313}, University of Illinois at Urbana–Champaign.

\bibitem{yates2014probability}
R.~D. Yates and D.~J. Goodman, \emph{Probability and stochastic processes: a friendly introduction for electrical and computer engineers}.\hskip 1em plus 0.5em minus 0.4em\relax John Wiley \& Sons, 2014.

\bibitem{weisstein2002euler}
E.~W. Weisstein, ``{Euler-Mascheroni} constant,'' \emph{https://mathworld. wolfram. com/}, 2002.

\bibitem{jiang2023mistral7b}
\BIBentryALTinterwordspacing
A.~Q. Jiang, A.~Sablayrolles, A.~Mensch, C.~Bamford, D.~S. Chaplot, D.~de~las Casas, F.~Bressand, G.~Lengyel, G.~Lample, L.~Saulnier, L.~R. Lavaud, M.-A. Lachaux, P.~Stock, T.~L. Scao, T.~Lavril, T.~Wang, T.~Lacroix, and W.~E. Sayed, ``Mistral 7b,'' 2023. [Online]. Available: \url{https://arxiv.org/abs/2310.06825}
\BIBentrySTDinterwordspacing

\bibitem{gemmateam2025gemma3technicalreport}
\BIBentryALTinterwordspacing
G.~Team, A.~Kamath, J.~Ferret, S.~Pathak, and et~al., ``Gemma 3 technical report,'' 2025. [Online]. Available: \url{https://arxiv.org/abs/2503.19786}
\BIBentrySTDinterwordspacing

\bibitem{abdin2024phi4technicalreport}
\BIBentryALTinterwordspacing
M.~Abdin, J.~Aneja, H.~Behl, S.~Bubeck, and et~al., ``Phi-4 technical report,'' 2024. [Online]. Available: \url{https://arxiv.org/abs/2412.08905}
\BIBentrySTDinterwordspacing

\bibitem{chung2022scalinginstructionfinetunedlanguagemodels}
\BIBentryALTinterwordspacing
H.~W. Chung, L.~Hou, S.~Longpre, B.~Zoph, Y.~Tay, and et~al., ``Scaling instruction-finetuned language models,'' 2022. [Online]. Available: \url{https://arxiv.org/abs/2210.11416}
\BIBentrySTDinterwordspacing

\bibitem{almazrouei2023falconseriesopenlanguage}
\BIBentryALTinterwordspacing
E.~Almazrouei, H.~Alobeidli, A.~Alshamsi, and et~al., ``The {Falcon} series of open language models,'' 2023. [Online]. Available: \url{https://arxiv.org/abs/2311.16867}
\BIBentrySTDinterwordspacing

\bibitem{joshi2017triviaqalargescaledistantly}
\BIBentryALTinterwordspacing
M.~Joshi, E.~Choi, D.~S. Weld, and L.~Zettlemoyer, ``{TriviaQA: A} large scale distantly supervised challenge dataset for reading comprehension,'' 2017. [Online]. Available: \url{https://arxiv.org/abs/1705.03551}
\BIBentrySTDinterwordspacing

\bibitem{clark2018thinksolvedquestionanswering}
\BIBentryALTinterwordspacing
P.~Clark, I.~Cowhey, O.~Etzioni, T.~Khot, A.~Sabharwal, C.~Schoenick, and O.~Tafjord, ``Think you have solved question answering? {Try ARC,} the {AI2} reasoning challenge,'' 2018. [Online]. Available: \url{https://arxiv.org/abs/1803.05457}
\BIBentrySTDinterwordspacing

\bibitem{talmor2019commonsenseqaquestionansweringchallenge}
\BIBentryALTinterwordspacing
A.~Talmor, J.~Herzig, N.~Lourie, and J.~Berant, ``{CommonsenseQA: A} question answering challenge targeting commonsense knowledge,'' 2019. [Online]. Available: \url{https://arxiv.org/abs/1811.00937}
\BIBentrySTDinterwordspacing

\bibitem{zheng2023judgingllmasajudgemtbenchchatbot}
\BIBentryALTinterwordspacing
L.~Zheng, W.-L. Chiang, Y.~Sheng, S.~Zhuang, Z.~Wu, Y.~Zhuang, Z.~Lin, Z.~Li, D.~Li, E.~P. Xing, H.~Zhang, J.~E. Gonzalez, and I.~Stoica, ``Judging {LLM-as-a-Judge} with {MT-Bench} and {Chatbot Arena},'' 2023. [Online]. Available: \url{https://arxiv.org/abs/2306.05685}
\BIBentrySTDinterwordspacing

\bibitem{chiang2023largelanguagemodelsalternative}
\BIBentryALTinterwordspacing
C.-H. Chiang and H.~yi~Lee, ``Can large language models be an alternative to human evaluations?'' 2023. [Online]. Available: \url{https://arxiv.org/abs/2305.01937}
\BIBentrySTDinterwordspacing

\end{thebibliography}

\end{document}